\documentclass[conference]{IEEEtran}
\IEEEoverridecommandlockouts
\usepackage{cite}
\usepackage{amsmath,amssymb,amsfonts}
\usepackage{algorithmic}
\usepackage{graphicx}
\usepackage{textcomp}
\usepackage{xcolor}
\def\BibTeX{{\rm B\kern-.05em{\sc i\kern-.025em b}\kern-.08em
    T\kern-.1667em\lower.7ex\hbox{E}\kern-.125emX}}

\newcommand{\plink}{ISCaP}
\newcommand{\SSD}{\name{PoPi}}
\newcommand{\ssd}{\mbox{\scriptsize{\name{PoPi}}}}
\newcommand{\SCA}{\name{SCA}}
\newcommand{\sca}{\mbox{\scriptsize{\name{SCA}}}}

\newcommand{\pob}{\mbox{\scriptsize{\name{Crowd}}}}

\newcommand{\poa}{\mbox{\scriptsize{\name{Nom}}}}
\newcommand{\FNA}{\mbox{\name{FNA}}}
\newcommand{\fna}{\mbox{\scriptsize{\name{FNA}}}}

\newcommand{\pc}{\mbox{$C$}}
\newcommand{\PC}{\mbox{$P_C$}}
\newcommand{\dc}{\mbox{$D_C$}}
\newcommand{\dt}{\mbox{$D_{\scriptsize{\name{MCrash}}}$}}

\newcommand{\ctx}[1]{\mbox{\textbf{\textup{C-#1}}}}
\newcommand{\mdl}[1]{\mbox{\textbf{\textup{M-#1}}}}
\newcommand{\clm}[1]{\mbox{\textbf{\textup{G-#1}}}}
\newcommand{\stg}[1]{\mbox{\textbf{\textup{S-#1}}}}
\newcommand{\sol}[1]{\mbox{\textbf{\textup{Sn-#1}}}}
\newcommand{\ass}[1]{\mbox{\textbf{\textup{A-#1}}}}

\newcommand{\name}[1]{\mbox{$\mathtt{#1}$}}

\newcommand{\tuple}[1]{\langle #1 \rangle}

\newtheorem{mymethod}{Method}
\newtheorem{mydefinition}{Definition}
\newtheorem{myobservation}{Observation}
\newtheorem{myremark}{Remark}
\newtheorem{myproposition}{Proposition}
\newtheorem{myclaim}{Claim}
\newtheorem{mylemma}{Lemma}
\newtheorem{mycorollary}{Corollary}
\newtheorem{myexample}{Example}
\newtheorem{myexamples}{Examples}
\newtheorem{myalgorithm}{Algorithm}
\newtheorem{myconstruction}{Construction}
\newtheorem{myrule}{Rule}

\newcommand{\bolddot}{\hspace{-1.5mm}\textbf{.}\ \  }

\newcommand{\BT}{\begin{theorem}}
\newcommand{\ET}{\end{theorem}}
\newcommand{\BCR}{\begin{mycorollary}\bolddot}
\newcommand{\ECR}{\end{mycorollary}}
\newcommand{\BPR}{\begin{myproposition}}
\newcommand{\EPR}{\end{myproposition}}
\newcommand{\BL}{\begin{mylemma}}
\newcommand{\EL}{\end{mylemma}}
\newcommand{\BCM}{\begin{myclaim}}
\newcommand{\ECM}{\end{myclaim}}

\newcommand{\BD}{\begin{mydefinition}}
\newcommand{\ED}{\end{mydefinition}}
\newcommand{\BPF}{\begin{proof}}
\newcommand{\EPF}{\end{proof}}
\newcommand{\BEX}{\begin{myexample}}
\newcommand{\EEX}{\end{myexample}}
\newcommand{\BEXS}{\begin{myexamples}}
\newcommand{\EEXS}{\end{myexamples}}
\newcommand{\BOB}{\begin{myobservation}}
\newcommand{\EOB}{\end{myobservation}}
\newcommand{\BR}{\begin{myremark}}
\newcommand{\ER}{\end{myremark}}
\newcommand{\BAL}{\begin{myalgorithm}}
\newcommand{\EAL}{\end{myalgorithm}}
\newcommand{\BAM}{\begin{mymethod}}
\newcommand{\EAM}{\end{mymethod}}

\newcommand{\BCO}{\begin{myconstruction}}
\newcommand{\ECO}{\end{myconstruction}}
\newcommand{\BRule}{\begin{myrule}}
\newcommand{\ERule}{\end{myrule}}

\makeatletter
\newcommand{\linebreakand}{%
  \end{@IEEEauthorhalign}
  \hfill\mbox{}\par
  \mbox{}\hfill\begin{@IEEEauthorhalign}
}
\makeatother

\begin{document}

\title{The missing link: Developing a safety case for perception components in automated driving \thanks{A journal version of this paper is available~\cite{sae2022-01-0818}.}}


\author{
\IEEEauthorblockN{Rick Salay, Krzysztof Czarnecki}
\IEEEauthorblockA{\textit{University of Waterloo}\\
Waterloo, Canada\\
\{rsalay, kczarnec\}@gsd.uwaterloo.ca}
\and
\IEEEauthorblockN{Hiroshi Kuwajima, Hirotoshi Yasuoka, Toshihiro Nakae}
\IEEEauthorblockA{\textit{DENSO CORPORATION} \\
Tokyo, Japan \\
\{hiroshi.kuwajima.j7d, hirotoshi.yasuoka.j2z, toshihiro.nakae.j8z\}@jp.denso.com}
\linebreakand
\IEEEauthorblockN{Vahdat Abdelzad, Chengjie Huang, Maximilian Kahn, Van Duong Nguyen}
\IEEEauthorblockA{\textit{University of Waterloo}\\
Waterloo, Canada\\
\{vahdat.abdelzad, c.huang, maximilian.kahn, harry.nguyen\}@uwaterloo.ca}

}

\maketitle
\thispagestyle{plain}
\pagestyle{plain}

\begin{abstract}
Safety assurance is a central concern for the development and societal acceptance of automated driving (AD) systems. Perception is a key aspect of AD that relies heavily on Machine Learning (ML). Despite the known challenges with the safety assurance of ML-based components, proposals have recently emerged for unit-level safety cases addressing these components. Unfortunately, AD safety cases express safety requirements at the system level and these efforts are missing the critical linking argument needed to integrate safety requirements at the system level with component performance requirements at the unit level. In this paper, we propose the Integration Safety Case for Perception (\textbf{\plink}), a generic template for such a linking safety argument specifically tailored for perception components. The template takes a deductive and formal approach to define strong traceability between levels. We demonstrate the applicability of \plink~with a detailed case study and discuss its use as a tool to support incremental development of perception components. 
\end{abstract}





\section{Introduction}
Safety assurance is a central concern for the development and societal acceptance of Automated Driving Systems (ADS). It is no coincidence that major players in this field have made their safety strategies publicly available (e.g.,~\cite{wood2019safety, webb2020waymo}). 
An ADS relies heavily on complex perception tasks to accurately determine the state of the world. These include image classification, object detection, and image segmentation using camera, LiDAR and Radar sensor data. Because these tasks are difficult to specify, machine learning (ML) is a preferred method of implementation; however, ML poses significant obstacles to safety assurance~\cite{salay18}. Despite this, the safety critical nature of perception requires reliable approaches to assuring their safety.

   
    

An ADS safety case aims to provide a hierarchical evidence-based argument for the claim that the ADS is acceptably safe. An important quality of a safety argument is ``rigor'', but when the steps of the argument are based on informal or non-deductive reasoning, this may be difficult to ensure. To address this, Rushby~\cite{rushby2015interpretation} has suggested that the internal decomposition steps of a safety case should be deductive, while inductive reasoning (e.g., generalizing from test results) are limited to the leaf claims that are supported directly by evidence. This is the approach we take in this paper to define a generic argument template for a perception component within an ADS. 

A limited amount of related work on safety cases in the ADS domain exists both at the whole system ADS level and at the unit-level for individual ML components. Kurd et al.~\cite{kurd2007developing} give a safety case for neural networks, and more recently, Burton et al.~\cite{burton2017making} give one for ML components in automated driving. Both use Goal Structuring Notation (GSN)~\cite{assurance2021goal}, but remain high-level. Wozniak et al.~\cite{wozniak2020safety} define a GSN argument pattern for ML components to produce an ISO 26262 style safety case. The pattern covers the refinement of unit-level safety requirements, data appropriateness, adequacy of the component design, and component training. The latter three areas are given a more detailed treatment with subclaims proposed.

Picardi et al.~\cite{picardi2020assurance} also focus on the unit level and sketch a safety case pattern in which each ML assurance claim is supported by a series of \emph{confidence} arguments that show why the claim is supported by context artifacts such as the test dataset, learned model, ML safety requirements, etc. These confidence arguments, in turn, draw on the ML development lifecycle~\cite{ashmore2019assuring} that specifies high-level requirements regarding these artifacts. For example, it specifies that test data should be ``relevant, complete, accurate and balanced''. The authors note that the ML assurance claims are drawn from ML safety requirements, which in turn are refined from system-level safety requirements based on methods such as hazard analysis; however, they have given no details on how this refinement is done.
As an illustration of how this could work, Gauerhof et al.~\cite{gauerhof2020assuring} study the elicitation of safety requirements for an ML-based pedestrian detector. For example, an analysis of the system-level safety requirement ``Ego shall stop at the crossing if a pedestrian is crossing'' yields several object detection component safety requirements such as ``Position of pedestrians shall be determined within 50\,cm of actual position''. Then the ML development lifecycle is used to guide the identification of detailed requirements for the confidence arguments---e.g., ``The data samples shall include sufficient range of environmental factors within
the scope of the ODD'' is a requirement to address test data completeness. Although this illustration is compelling, it offers no general approach for connecting system-level to component unit-level safety requirements. 

The recent work by Bloomfield et al.~\cite{bloomfield2021safety} is an extensive effort to define a safety case template using the Claims-Argument-Evidence (CAE) notation, for autonomous systems that include ML components. It assumes that ML components will be coupled with monitors that guard the component against bad behaviour.  The template focuses on the adequacy of hazard analysis at the system level and gives high-level templates for arguments at the monitor+ML subsystem level as well as the unit level for the ML component. The reasoning approach advocated is informal with an emphasis on identifying potential \emph{defeaters} that challenge claims. As with the work of Picardi et al.~\cite{picardi2020assurance}, there is a brief discussion about connecting the system-level requirements to the unit-level but details are missing. For example, ``The number of crashes involving the AV averages at most 89 crashes per million miles driven with confidence 95\%.'' is given as a sample system-level claim and ``YOLOv3 correctly identifies traffic lights in 87\% of images containing traffic lights'' is a claim at the unit-level but the parts of the argument that show how the performance of object detector YOLOv3 impacts the crash rate of the AV are not addressed by the template.

The Aurora safety case framework~\cite{aurora21} is another recent effort defining a safety case structure addressing the entire ADS development process. It consists of a GSN decomposition tree of generic claims from the top claim that the ADS is acceptably safe. Although its scope is broad, its claims remain at high-level, are non-formal, and do not refine to component-level claims. For example, the leaf claim ``The systems engineering process is appropriate for safety critical design'' is typical of the most refined level of detail in the framework.

\begin{figure}
    \centering
    \includegraphics[width=0.45\textwidth]{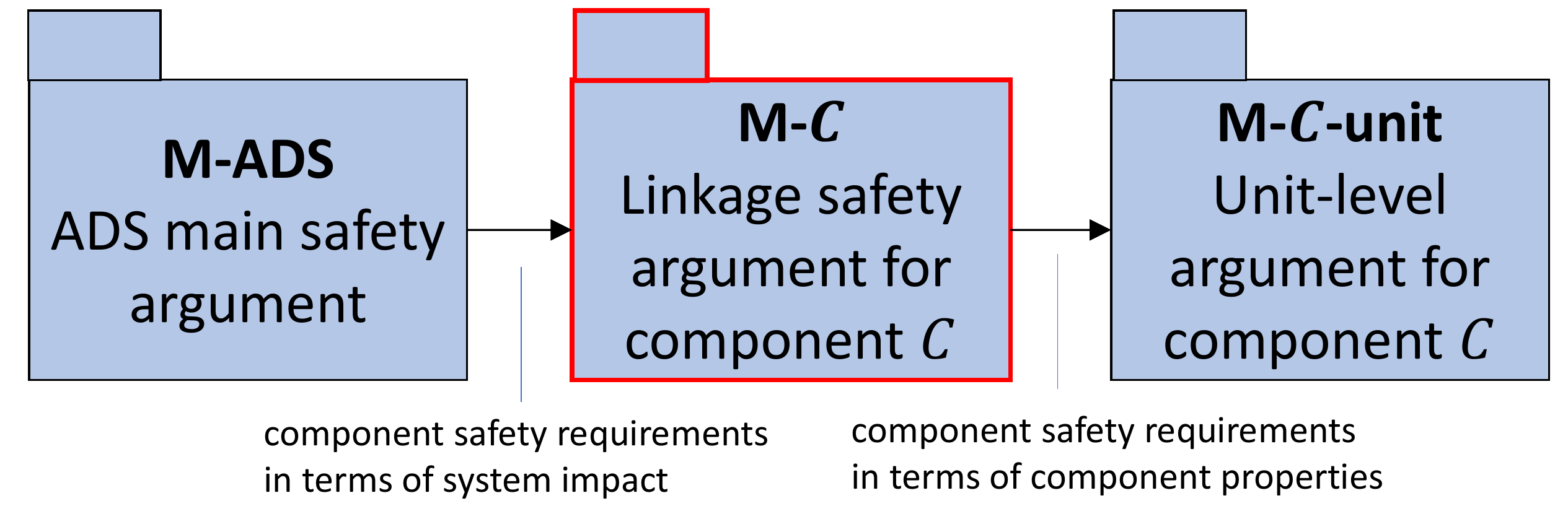}
    \caption{Role of the linking argument M-$C$ instantiated from \plink~presented in this paper, where \pc~represents a perception component. Arrows represent the ``is supported by'' relation where a higher-level argument is supported by a lower-level one.}
    \label{fig:positioning}
\end{figure}

In contrast to the work discussed above, in this paper we focus specifically on addressing the connection between the system-level and unit-level arguments. The recent work of  Vaicenavicius et al.~\cite{vaicenavicius2020self} studies this connection formally for a specific simple driving scenario and sketches the safety case structure for this scenario. While this work is in the same spirit as ours, it is limited to a simple idealized illustration and does not consider the special characteristics of perception components such as vulnerability to environmental conditions affecting perception. In this paper, we propose the Integration Safety Case for Perception (\textbf{\plink}, pronounced Ice-Cap), a template that provides a systematic method of generating a formal ``linkage argument'' between the system-level and unit-level arguments, specifically for perception components. This is illustrated in Fig.~\ref{fig:positioning} using GSN safety case modules. The safety case module \mdl{$C$} instantiated from \plink~for a perception component \pc, sits between the main ADS argument module \mdl{ADS} that relies on a claim about component contribution to system safety and the unit-level argument module \mdl{$C$-unit} that provides evidence for specific component properties required for safety. 

Our contributions are:
    \begin{itemize}
        \item \plink, a formal safety case template linking system and unit arguments using a safety claim decomposition method based on \emph{hazardous misperception patterns} that separates analysis of ADS dynamics from perception.
        \item The use of \plink~to identify a set of \emph{risk-aware} performance metrics that are tailored to the perception component and the ADS in which it operates. 
        \item A safety case structure that has desirable properties including stability with respect to component changes and support for assessing risk trade-offs in different operational situations.
    \end{itemize}
    
The remainder of the paper is structured as follows. In Sec~\ref{sec:prelim} we give the formalization preliminaries. Sec.~\ref{sec:safetycase} presents \plink~using GSN notation followed by a discussion in Sec.~\ref{sec:discussion} on the application of \plink~in component development. The use of \plink~is illustrated with a detailed example in Sec.~\ref{sec:casestudy}. Finally, in Sec.~\ref{sec:conclusion} we give conclusions and discuss future work.

\section{Preliminaries}\label{sec:prelim}
In this section, we define the terms and notation used in \plink.
\subsection{Perception Tasks}
We distinguish between the perception task $T$ in the ADS and a component \pc~that implements the task. For example, $T$ may represent the task ``detect and localize vehicles surrounding the ego vehicle'' while \pc$_1$ and \pc$_2$ may be alternate components that implement this task with different performance characteristics. 

\BD[Perception task]\label{def:ptask}
A \emph{perception task} $T$ is represented as function $T:X\to Y$ from input domain $X$ to output codomain $Y$  defining the ``ground truth'' of how the task should be performed.
If component $C$ implements $T$, then it defines function $C:X\to Y$ intended to approximate $T:X\to Y$. 
The unit of perception for $T$, called a \emph{frame}, is denoted $(x,y)\in X\times Y$. 
The \emph{frame rate} is the number of frames per unit time. The distribution $P_{C,\textit{fr}}(x,y)$ denotes the probability of frame $(x,y)$ occurring while the ADS operates a vehicle in the ODD using component $C$ that implements $T$. 
\ED

For example, task $\name{OD}$ for camera-based object detection defines function $\name{OD}:\name{CImage}\to\name{BBSet}$ from camera images to bounding box sets. Object detector $\name{Yolo}:\name{CImage}\to\name{BBSet}$ implements $\name{OD}$.
A frame is a single camera image and corresponding output set of bounding boxes.

\subsection{Drives}
We define notation for describing driving scenarios.

\BD[States]
A state $s$ is a snapshot of all relevant ADS and environment state variables at a point in time. $\mathcal{S}$ is the set of possible states.
\ED

The rate at which snapshots are taken determines how fine-grained in time driving is represented. For the argument \mdl{\pc}, it is convenient to take a task-centric view and assume that this is the frame rate for $T$. 

ADS operation can be viewed as an infinite stochastic process producing a sequence of states. We can classify the states that occur during ADS operation using temporal properties.
\BD[State classification]
Given state $s_t$, $s\in \mathcal{S}, t\in \mathbb{Z}$, 
$\PC(s_t\in \phi)$ (or $\PC(\phi)$) denotes the probability that a randomly chosen state during ADS operation in the ODD using component \pc~satisfies temporal property $\phi$. $\name{MCrash}$ denotes the temporal property identifying crash states caused by some preceding sequence of misperceptions in performing task $T$. 
\ED 

\vspace{0.1in}

Thus, $\PC(\name{MCrash})$ denotes the probability that a  misperception-caused crash state occurs when using component \pc.

\BD[Drives]
A \emph{drive} is a finite sequence $d\in\mathcal{S}^*$ of states. $d'\subseteq d$ denotes that drive $d'$ is a subsequence of $d$. \name{Matches}$(d,s_t)$ is a predicate that holds iff $\forall k\in \{0,...,n-1\}\cdot s_{t-k}=d[n-k]$,  where, $n$ is the length of $d$ and $d[i]$ is the $i^{th}$ state in $d$.

\ED

Thus, \name{Matches}$(d,s_t)$ means that $s_t$ and its $n-1$ preceding states match $d$. 

\subsection{Misperceptions}
The safety of perception tasks is impacted by the presence of misperceptions.
\BD[Misperception] \label{def:misperc}
Given task $T$, a \emph{misperception} is a frame $(x,y)$ such that $y\neq T(x)$. A misperception \emph{by component} $C$ is a frame $(x, C(x))$ that is a misperception. 
\ED

The safety impact of a misperception varies depending on the context in which it occurs. For example, in a vehicle detection task, a false negative (FN) (i.e., not detecting a vehicle) or false positive (FP) (i.e., falsely detecting a vehicle) close to the ego vehicle  may be hazardous, but when these occur far away, they may be benign. Although having too many benign misperceptions can negatively impact ADS performance, in the safety argument for component $C$  we consider only the hazardous misperceptions it can produce. 
In order to characterize misperceptions, we generalize from individual misperceptions to \emph{patterns} of misperceptions. 

\BD[Misperception Pattern] \label{def:mispat}
A \emph{Misperception Pattern (MP)} identifies a subset of drives in which every drive contains some states with misperceptions. 
A \emph{Frame Misperception Pattern} applies to a single state and is defined as function $\name{fMP}:Y\to pow(Y)$ such that $\forall y\in Y\cdot y\notin \name{fMP}(y)$, where $pow(Y)$ is the power set of $Y$. 
State $s$ containing frame $(x,y)$ \emph{satisfies} frame misperception pattern \name{fMP} 
iff  $y\in \name{fMP}(T(x))$.  
\ED

Misperception patterns can be used to categorize misperceptions based on conditions of interest. For example, the frame misperception pattern \name{FN_{20}} can denote all object detection misperceptions that include false negatives within $20\,m$ of the ego vehicle.  Note that  for some values of $T(x)\in Y$   there may be no instances of the frame misperception pattern $\name{FN_{20}}(T(x))$.  For example, if $T(x)$ contains no detections within $20\,m$ of the ego vehicle then \name{FN_{20}}$(T(x))=\emptyset$ since there  is no way to get such an FN. The misperception pattern \name{30FN_{20}} can denote the set of all drives satisfying \name{FN_{20}} in at least 30\% of its states. Thus,  misperception patterns are naturally defined in terms of frame misperception patterns and we exploit this in the safety argument.

\subsection{Hazardous Misperceptions}
Following ISO 26262~\cite{ISO26262} and SOTIF~\cite{ISO21448}, system-level hazard analysis identifies cases where a driving scenario combined with a hazardous behaviour by the ego vehicle and particular reactions by scenario participants will result in harm (i.e., a crash). For example, the ego vehicle waiting to turn left at an intersection is a scenario in which, if the ego vehicle begins turning with an on-coming vehicle too close (hazardous behaviour), and the on-coming car cannot stop (participant reaction), a collision will occur. We define a term to represent these cases.
\BD[Hazardous Behaviour Sensitive Scenario]
A \emph{Hazardous Behaviour Sensitive Scenario (HBSS)} is a subset of drives exhibiting a particular combination of operational scenarios with participant reactions in which there are possible hazardous ego vehicle behaviours.   
\ED

In some cases, the hazardous behaviour in an HBSS can be \emph{caused by} a sequence of frame misperceptions in performing perception task $T$. For example, in the HBSS with the ego vehicle turning left, the hazardous behaviour could be caused by a sequence of misperceptions in the \name{OD} task over several frames that leads the ADS to believe there is no vehicle coming and allows a hazardous left turn to be performed. We refer to such sequences as \emph{hazardous misperceptions} and formalize their occurrence as a type of misperception pattern.

\BD[Hazardous Misperception Pattern]\label{def:hmp}
 A \emph{hazardous misperception pattern (HMP)} is a pair $\tuple{\name{HBSS},\name{MP}}$ of predicates over set $\mathcal{S}^*$ of drives where,  
  
 \begin{itemize}
     \item \name{HBSS} is the \emph{Hazardous Behaviour Sensitive Scenario condition} that specifies properties of the drive required for the scenario to occur and no other drive factors. At most one state in the drive can be a crash state and it must occur at the end of the drive.
     \item \name{MP} is the \emph{Misperception Pattern condition} that specifies \emph{all} sequences of misperceptions that will cause a hazardous behaviour in drives that satisfy the \name{HBSS} condition. The \name{MP} condition constrains only drive factors essential for describing the misperceptions.

 \end{itemize}

A drive $d\in \mathcal{S}^*$ \emph{satisfies} the HMP iff 
$$\name{HBSS}(d)\wedge (\exists d'\in \mathcal{S}^*\cdot \name{MP}(d')\wedge d'\subseteq d)$$ 
\ED

The condition \name{HBSS} says that the HBSS for the system hazard associated with HMP occurs over the length of the drive.
Thus, a drive satisfying \name{HBSS} represents a complete scenario that ends in a crash if any hazardous behaviour by the ego vehicle, as identified by the HBSS, occurs.  

Condition \name{MP} identifies the sequences of misperceptions performing task $T$ that, when they occur within an \name{HBSS} drive, will cause a hazardous behaviour to occur leading to a crash. 
This condition would naturally be expressed in terms of various frame misperception patterns. 

For example, assume HMP \name{IL} is based on the  $\name{HBSS}_{\name{IL}}$ condition that identifies a drive in which the ego vehicle is waiting to turn left at an intersection. We focus on the hazardous behaviour in which the ego vehicle begins turning with an on-coming vehicle too close, causing a collision. This hazardous behaviour could be caused by misjudging the position and/or speed of the on-coming vehicle (mis-localization) or not detecting the on-coming vehicle (FN).
With analysis and experimentation, it is possible to determine the exact sequences of these misperceptions that would cause the hazardous behaviour and these are used to define $\name{MP}_{\name{IL}}$.

Although drive predicates such as \name{HBSS} and \name{MP} play an important role in the safety argument, we do not specify a formal language for expressing drive predicates and instead remain at the semantic level, thinking of predicates in terms of the sets they define. For example, drive predicates can be defined using a general language such as First Order Logic (FOL), or more specialized logics such as temporal logic (e.g., Linear Temporal Logic). Remaining ``syntax agnostic'' gives \plink~the flexibility to be used in different analytical contexts.

\section{The \plink~Safety Case Template} \label{sec:safetycase}

\begin{figure*}
    \centering
    \includegraphics[width=1\textwidth]{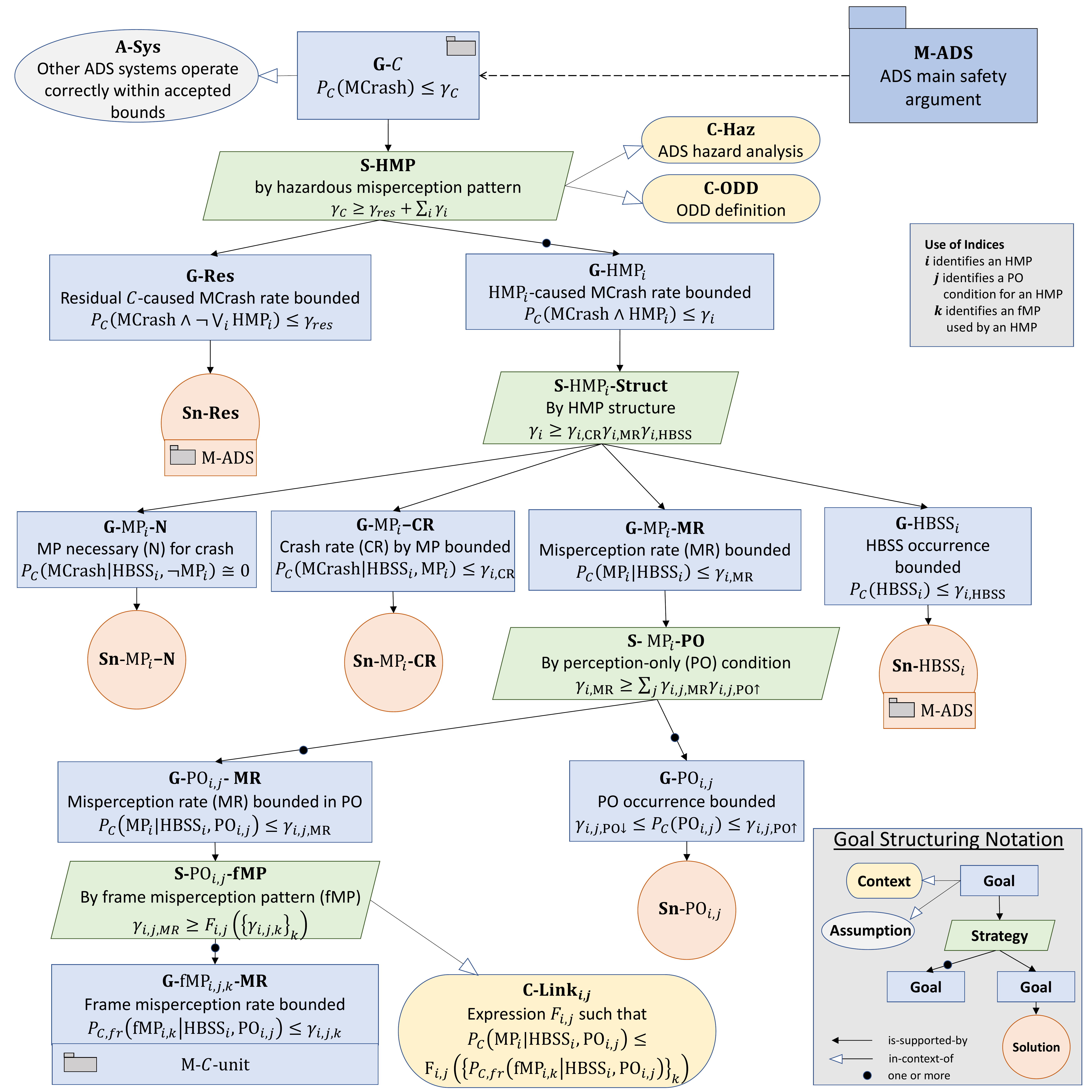}
    \caption{The \plink~safety case template for the argument in \mdl{\pc} using GSN.}
    \label{fig:sct}
\end{figure*}

In this section, we describe in detail, the claims, decomposition strategies, and sources of evidence for the argument in \mdl{\pc} as an instantiation of the \plink~safety case template.  Fig.~\ref{fig:sct} shows the rendering of \plink~in GSN. Claims are expressed using \emph{Goal} nodes, decompositions of claims into sub-claims as \emph{Strategy} nodes and evidence as \emph{Solution} nodes. In addition, \emph{Context} and \emph{Assumption} nodes denote supporting information. 

We exploit the modularity and templating features in GSN to encapsulate the argument instantiated from \plink~in a separate module \mdl{\pc} and treat the top-level goal \clm{\pc} as an \emph{away-goal}---i.e., referenced from within the main ADS argument \mdl{ADS}. While details of \mdl{ADS} and \mdl{\pc-unit} are left unspecified and are out of scope for this paper, we highlight aspects of them that \mdl{\pc} depends on.  

Overall, the argument takes a formal deductive approach in which claims are expressed as bounded probabilities and strategies mathematically relate the bounds between child and parent claims. This restricts all inductive steps of the argument to the solution nodes that provide evidence for leaf claims.
The bound in the top claim for component \pc~is first decomposed using HMPs corresponding to system-level hazards that have misperceptions as causes. The bound in each claim for a specific HMP is then factorized into claims that bound the crash rate, misperception pattern occurrence rate, and HBSS (i.e.,  hazardous ADS dynamics) occurrence rate, plus a guarantee that all hazardous misperceptions in the HBSS are covered. Among these claims, the bound on the misperception pattern rate is further decomposed using perception-only (PO) conditions, separating the misperception rate for each PO condition and rate of occurrence for the PO condition. Finally, the bound on the misperception pattern rate for each PO condition is decomposed into bounds on the occurrence rates of the constituent frame misperception patterns. These define risk-aware performance metrics that can be estimated via testing in the machine learning context.

\subsection{Goal \clm{\pc}}
\plink~assumes a single type of high severity event, which we term ``crash'' throughout this paper. In Sec.~\ref{sec:severity}, we discuss how to adapt \plink~to allow for multiple severity levels.

The top-level claim, that component \pc~is adequately safe, is formalized by bounding the probability \PC$(\name{MCrash})$ that the ADS reaches a crash state due to a misperception by component \pc~in performing task $T$. Note that we take an assume-guarantee view of \mdl{\pc}, where claim \clm{\pc} is guaranteed (up to the evidence) only if we assume all other ADS components are operating as designed (\ass{Sys}). This assumption can be adjusted to accommodate different system considerations. For example, if \pc~is a camera-based object detector, we could assume a particular failure rate of the camera. This would have to then be incorporated into the claims since then sometimes \name{MCrash} may occur due to camera failure rather than a misperception of the object detector.

\begin{figure}[!t]
    \centering
    \includegraphics[width=0.40\textwidth]{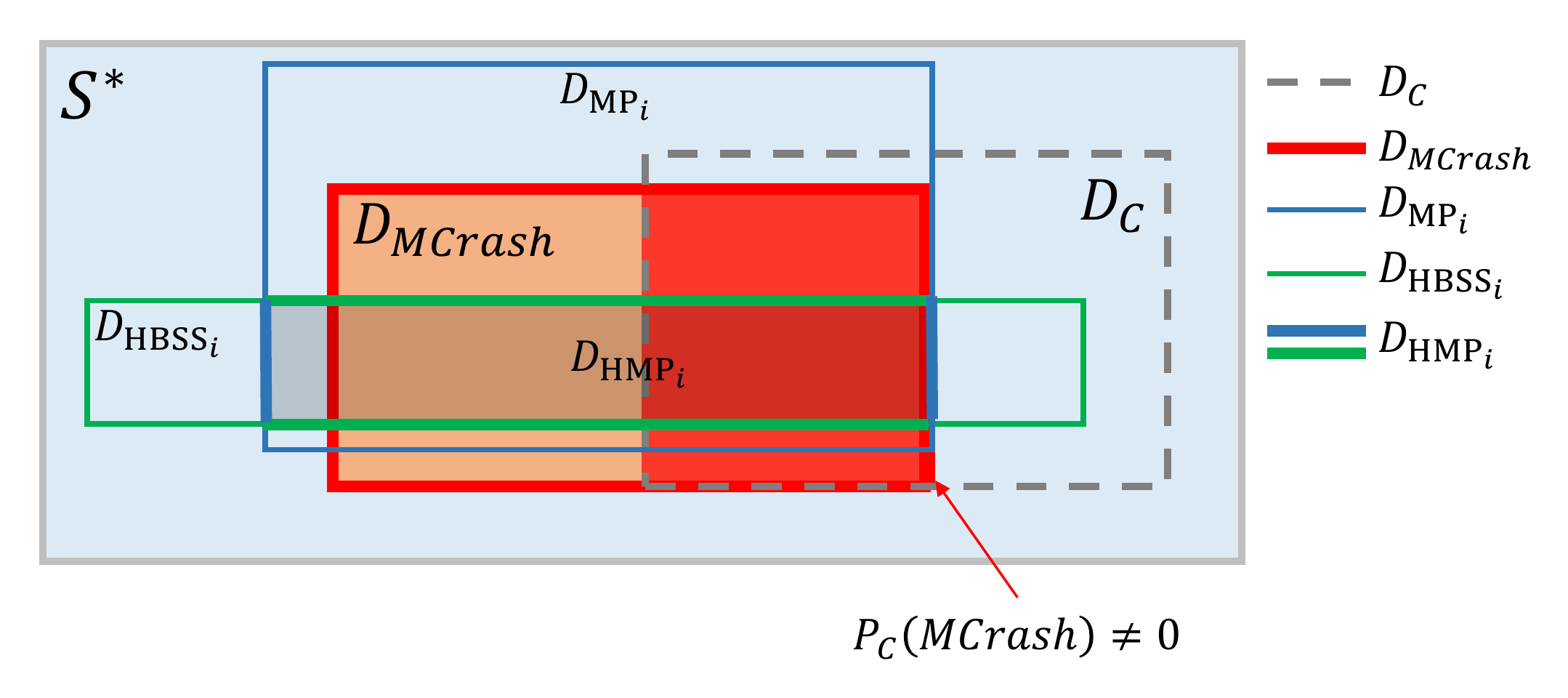}
    \vspace{0in}
    \caption{Decomposition of \clm{\pc} into sub-claims by decomposing \dt~into subsets corresponding to hazardous misperception patterns.}
    
    \label{fig:s-mp}
\end{figure}

Fig.~\ref{fig:s-mp} helps explain the meaning of \clm{\pc}. 
Subset \dt~are drives that contain hazardous misperceptions in carrying out task $T$. Subset \dc~are drives that are possible when component \pc~implements $T$. Component \pc~may not produce some hazardous misperceptions, and other misperceptions it produces may be benign; thus, it is the occurrence of drives in the intersection \dt$\cap$\dc~(bright red) that the claim \clm{\pc} seeks to bound to an acceptable level. Safer components have a smaller value for \PC$(\name{MCrash})$, and if the current component \pc~does not meet the required target level, different mitigation steps could be taken. For example, \pc~could be replaced with a better component \pc$'$ that makes fewer hazardous misperceptions, an ensemble of components can be used jointly to reduce hazardous misperceptions using redundancy, etc. 

\subsection{Strategy \stg{HMP}}
It is unrealistic to expect to provide evidence directly for \clm{\pc} since this covers all possible misperception-caused crashes. The similar complexity issue in the main ADS argument is addressed by decomposing the safety claim based on system hazards that may occur. For a system hazard that can be caused by hazardous misperceptions, we can define a corresponding HMP. 

The strategy \stg{HMP} exploits this use of system hazards to decompose \dt~ and, correspondingly, \clm{\pc} into sub-claims using HMPs.  
In Fig.~\ref{fig:sct}, \stg{HMP} is linked to context information about the ADS hazard analysis and creates sub-claims \clm{\name{HMP}$_i$} for the corresponding identified hazardous misperception patterns. In addition, it includes the sub-claim \clm{Res}, which corresponds to the residual hazardous misperception-caused drives not covered by any HMP. 
\begin{figure}
    \centering
    \includegraphics[width=0.44\textwidth]{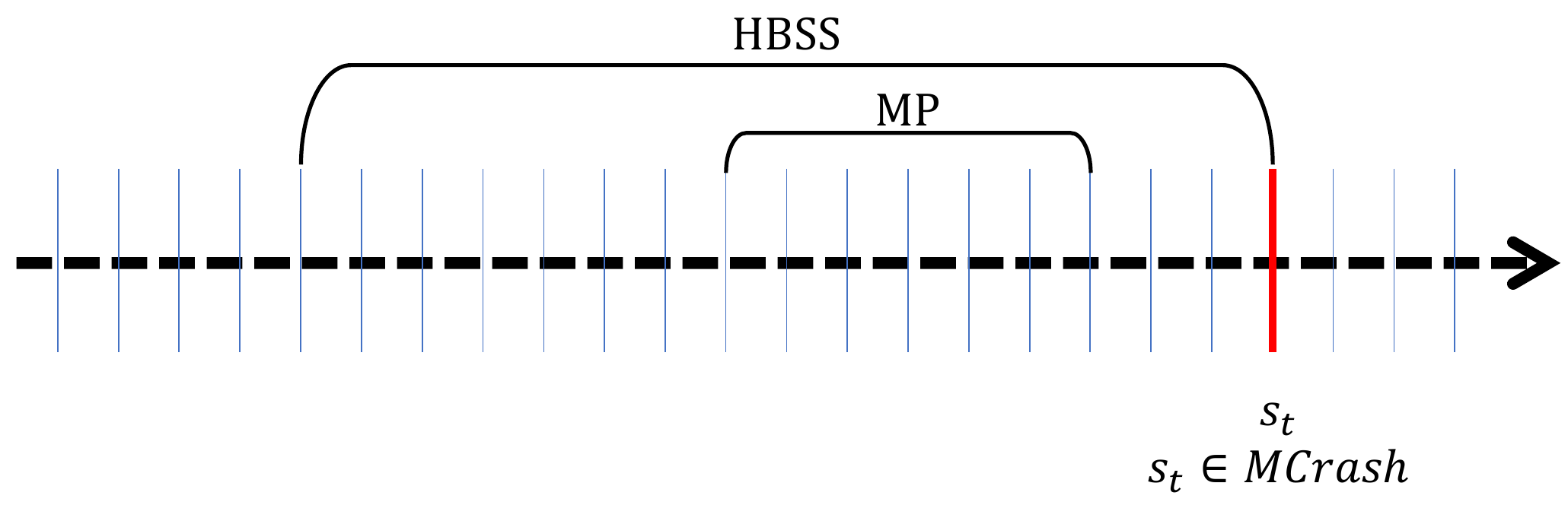}
    \caption{Illustration of how an HMP can be used to classify cases of misperception-caused crashes.}
    \label{fig:hmp}
\end{figure}

To simplify the relationship between bounds, we assume that the HMPs are chosen so that the subsets of drives they define are disjoint.  We show below that it is always possible to satisfy this restriction. With this, the simple additive relationship  $\gamma_{\scriptsize{\pc}}\ge\gamma_{res}+\Sigma_i\gamma_i$ holds among the sub-claims. This allows the bound on \PC$(\name{MCrash})$ to be conveniently seen as a ``risk budget'' that is allocated to different HMPs and helps to study risk trade-offs. 
Mathematically, this strategy can be understood as an application of the law of total probability: $\PC(\name{MCrash})=\PC(\name{MCrash}\wedge \neg \bigvee_{i} \name{HMP}_{i})+\bigvee_{i}\PC(\name{MCrash}\wedge \name{HMP}_i)$. 

\subsection{Goal \clm{\name{HMP}$_i$}}

Def.~\ref{def:hmp} defines an HMP as a drive specification in terms of HBSS and MP conditions.  Fig.~\ref{fig:s-mp} shows an example subset of drives $D_{\text{\name{HMP}}_i}$ that satisfy the specification for one \name{HMP}$_i$. The corresponding sub-claim \clm{\name{HMP}$_i$} bounds the occurrence of drives in the subset $D_{\text{\name{HMP}}_i}\cap$\dc~--- i.e. the drives in {\name{HMP}}$_i$ that could be produced when component \pc~is used. As discussed above, we assume that HMPs identify disjoint sets of drives. This is always possible to achieve by constructing additional HMPs corresponding to overlaps. For example, if \name{HMP}$_i$ and \name{HMP}$_j$ overlap, we define \name{HMP}$_{i,j} \equiv\tuple{\name{HBSS}_i \wedge \name{HBSS}_j, \name{MP}_i \vee \name{MP}_j}$ and redefine \name{HMP}$_{i} \equiv \tuple{\name{HBSS}_i \wedge \neg \name{HBSS}_j, \name{MP}_i}$ and \name{HMP}$_{j} \equiv \tuple{\neg\name{HBSS}_i \wedge \name{HBSS}_j, \name{MP}_j}$.


 \subsection{Goal \clm{Res} and Solution \sol{Res}}
  Claim \clm{Res} bounds the occurrence of hazardous misperceptions in drives not covered by any HMP---those in the residual subset $(\dt\cap\dc)\setminus\bigcup_i D_{\text{\name{HMP}}_i}$ in Fig.~\ref{fig:s-mp}. Since, the HBSSs are based on system-level hazards, the evidence for claim \clm{Res}  should be drawn from evidence in argument \mdl{ADS} regarding completeness of hazard analysis. We identify \sol{Res} as an away-solution to indicate this fact. 
  
  For example, \mdl{ADS} may use a reference such as the NHTSA pre-crash scenario typology~\cite{najm2007pre} to provide evidence for sufficient coverage of hazard scenarios. These scenarios account for $99.4\%$ of all reported light-vehicle crashes and the HMPs could be based on the subset of these that can be caused by misperceptions. Thus, subject to the assumptions that data about human-caused crashes can be applied to ADS-caused crashes and that we are limited to reported light-vehicle crashes, the residual of $0.6\%$ could be used as a basis for residual bound $\gamma_{res}$.
 
 \subsection{Strategy \stg{\name{HMP}$_i$-Struct}}
 Fig.~\ref{fig:hmp} shows how an HMP is related to a misperception-caused crash state. The crash state $s_t$ occurs because the drive preceding it was an HBSS drive, and within this, an MP misperception sequence caused a hazardous behaviour. 
Given this fact and Def.~\ref{def:hmp}, we can express  $\PC(\name{MCrash}\wedge \name{HMP}_i)$ as
\begin{equation*}
\begin{split}
\PC(s_t\in\name{MCrash} & \wedge 
\exists d\in \mathcal{S}^*\cdot \name{HBSS}_i(d)\wedge \name{Matches}(d, s_t) \\ 
&\wedge \exists d'\in\mathcal{S}^*\cdot\name{MP}_i(d')\wedge d'\subseteq d)
\end{split}
\end{equation*}
This can be decomposed into the product of three terms using the chain rule for joint probabilities:


1. Misperception-caused crash rate 
\begin{equation*}
    \begin{split}
\PC(s_t\in\name{MCrash}&  |\exists d\in \mathcal{S}^*\cdot \name{HBSS}_i(d)\wedge \name{Matches}(d, s_t) \\ & \wedge \exists d'\in\mathcal{S}^*\cdot\name{MP}_i(d')\wedge d'\subseteq d)
    \end{split}
\end{equation*}
    \begin{itemize}
        \item Meaning: Probability that a state $s_t$ is a misperception-caused crash given it ends an $\name{HBSS}_i$ drive containing an $\name{MP}_i$ sequence.
        \item Shorthand: $\PC(\name{MCrash}|\name{HBSS}_i,\name{MP}_i)$
    \end{itemize} 
    
2. Misperception rate
\begin{equation*}
    \begin{split}
\PC(\exists d\in \mathcal{S}^*\cdot & \name{HBSS}_i(d)\wedge \name{Matches}(d, s_t) \\ & \wedge \exists d'\in\mathcal{S}^*\cdot\name{MP}_i(d')\wedge d'\subseteq d \\
&|\exists d\in\mathcal{S}^*\cdot\name{HBSS}_i(d) \wedge \name{Matches}(d, s_t))
    \end{split}
\end{equation*}
    \begin{itemize}
        \item Meaning: Given a state $s_t$ ends an $\name{HBSS}_i$ drive, the probability the state is preceded by an $\name{MP}_i$ sequence contained within  the drive. 
        \item Shorthand: $\PC(\name{MP}_i|\name{HBSS}_i)$
    \end{itemize}
    
3. Exposure of HBSS in ODD 
\begin{equation*}
    \PC(\exists d\in \mathcal{S}^*\cdot \name{HBSS}_i(d)\wedge \name{Matches}(d, s_t))
\end{equation*}

    \begin{itemize}
                \item Meaning: The probability that a state $s_t$ ends an $\name{HBSS}_i$ drive. 
        \item Shorthand: $\PC(\name{HBSS}_i)$
    \end{itemize}

These three terms produce the sub-claims \clm{\name{MP}$_i$-CR}, \clm{\name{MP}$_i$-MR} and \clm{\name{HBSS}$_i$}, respectively. Thus, we have that $\gamma_i \ge \gamma_{i,\scriptsize{\name{CR}}}\gamma_{i,\scriptsize{\name{MR}}}\gamma_{i,\scriptsize{\name{HBSS}}}$. In Fig.~\ref{fig:sct} the shorthand forms of the probabilities are used. The fourth sub-claim, \clm{\name{MP}$_i$-N}, is needed to show that $\name{MP}_i$ covers all hazardous misperceptions in $\name{HBSS}_i$.

\subsection{Goal \clm{\name{MP}$_i$-N} and Solution \sol{\name{MP}$_i$-N}}
By Def.~\ref{def:hmp}, $\name{MP}_i$ must identify all hazardous misperceptions that component \pc~could produce in an $\name{HBSS}_i$ drive---i.e., that \name{MP}$_i$ is \emph{necessary} for a misperception caused crash in this HBSS. In Fig.~\ref{fig:s-mp}, this is equivalent to the condition: 
$$(\dt\cap\dc\cap D_{\scriptsize{\name{HBSS}_i}})\subseteq (D_{\scriptsize{\name{MP}_i}}\cap\dc\cap D_{\scriptsize{\name{HBSS}_i}})$$ 
Claim \clm{\name{MP}$_i$-N} states this by saying that an $\name{HBSS}_i$ drive not containing an $\name{MP}_i$ misperception sequence has probability zero of ending in a crash.

Strong evidence for this claim may require safety analysis methods such as FTA and FMEA (or its specializations, e.g., CFMEA~\cite{salay2019safety}).
Note that this analysis would establish that the \name{MP}$_i$ is necessary for a \emph{hazardous behaviour to occur} under the \name{HBSS}$_i$ conditions. The analysis identifying the hazardous behaviours (i.e., behaviours that lead to a crash) under \name{HBSS}$_i$ is a separate analysis that should be done at the vehicle level as part of \mdl{ADS}.
Empirical evidence could be obtained by simulating randomly sampled \name{HBSS}$_i$ drives and checking that those that end in a crash also satisfy \name{MP}$_i$.

\subsection{Goal \clm{\name{MP}$_i$-CR} and Solution \sol{\name{MP}$_i$-CR}}\label{sec:mp-i-s}
The crash rate is the probability that a drive in \name{HMP}$_i$ ends in a crash. The  \name{HBSS}$_i$ condition alone only guarantees that a misperception-caused hazardous behaviour is \emph{possible} in the drive, but doesn't guarantee one occurs. The condition \name{MP}$_i$ in \name{HMP}$_i$ constrains \name{HBSS}$_i$ to drives that contain misperceptions. When \name{MP}$_i$ is sufficiently constraining so that the only misperceptions that satisfy it are hazardous (i.e., no benign misperceptions), then a crash is guaranteed, and thus $\PC(\name{MCrash}|\name{HBSS}_i,\name{MP}_i)=1$. This is desirable because allowing benign misperceptions into \name{HMP}$_i$ forces the bound $\gamma_{i,\scriptsize{\name{CR}}}$ to be less tight than necessary. However, allowing benign misperceptions does not pose a safety risk and may simplify analysis. One could also conservatively set $\gamma_{i,\scriptsize{\name{CR}}}=1$ as long as the resulting bound in the top claim is sufficiently low (this is done in the case study), otherwise a tighter bound may be needed.

Empirical evidence to check or estimate $\gamma_{i,\scriptsize{\name{CR}}}$ could be obtained by simulating randomly sampled \name{HBSS}$_i$ drives with randomly injected \name{MP}$_i$ misperceptions and counting the number that end in a crash.

\subsection{Goal \clm{\name{HBSS}$_i$} and Solution \sol{\name{HBSS}$_i$}}
The claim in this goal bounds the occurrence of \name{HBSS}$_i$ drives in the ODD. In the context of ISO 26262 and SOTIF, this is related to the level of  \emph{exposure}\footnote{Note that in the current version of \plink~we assume that exposure is  frequency-based. Additionally supporting duration-based exposure is left as future work.} to the \name{HBSS}$_i$ scenario by the ADS. 
 

Since the exposure to \name{HBSS}$_i$ is the same as the exposure to the corresponding system-level hazard it is based on, the exposure level and evidence should already be found in argument \mdl{ADS}. Thus, the \sol{HBSS$_i$} is identified as an away-solution node.

Although this claim uses distribution \PC~specific to component \pc, in practice, changing the component should not affect the exposure. The exposure level would only be dependent on a specific choice of \pc~if the ADS explicitly accounted for the weaknesses of \pc~in its driving policy. For example, if the object detector being used was known to perform poorly on busy roads, the ADS policy could be designed to avoid them, thus affecting exposure levels. 




\subsection{Goal \clm{\name{MP}$_i$-MR}} 
The misperception rate is the key subclaim of the decomposition \stg{\name{HMP}$_i$-Struct} that is affected by the behaviour of component \pc; thus, it is the focal point of any development effort to improve safety by changing \pc. However, estimating the bound $\gamma_{i,\text{MR}}$ directly may be difficult. Since the development goal for component \pc~is to make it perform well in the ODD, \name{MP}$_{i}$ misperceptions may not be produced by \pc~in the majority of \name{HBSS}$_{i}$ drives, making this a rare event and challenging for collecting empirical evidence. 
However, it is well known that perception performance can be dramatically affected by external conditions such as weather, lighting, object properties such as shape, color and size, spatial configurations of objects, etc. We refer to these as \emph{Perception-Only (PO)} conditions since they identify sub-cases within \name{HBSS}$_{i}$ with higher probability of \name{MP}$_{i}$ misperceptions but do not constrain ADS behaviour.
Distinguishing HBSS from MP and PO conditions allows a separation of analysis of ADS dynamics (i.e., planning and actuation) from perception.
In a similar spirit to hazard analysis at the higher levels of the safety case, we decompose the claim \clm{\name{MP}$_i$-MR} according to PO conditions.

\subsection{Strategy \stg{\name{MP}$_i$-PO}}
Let $\{\name{PO}_{i,j}\}_j$ be a set of PO conditions for \name{HMP}$_i$. PO conditions are drive predicates that further ``condition'' the drives within $D_{\scriptsize{\name{HBSS}_i}}$ in Fig.~\ref{fig:s-mp} by constraining drive factors that affect the occurrence of misperceptions. 
We assume that PO conditions define disjoint subsets of drives within $D_{\scriptsize{\name{HBSS}_i}}$. Finally, we define $\name{PO}_{i,\scriptsize{\name{Nom}}}=\neg\bigvee_{j\neq \scriptsize{\name{Nom}}}\name{PO}_{i,j}$ as the distinguished PO condition representing the nominal case of drives where no other PO condition holds. 
With this, the subsets $\{\name{PO}_{i,j}\}_j$ partition the set $D_{\scriptsize{\name{HBSS}_i}}$. 
Based on these assumptions and by the law of total probability, we have:
\begin{multline}\label{eqn:po}
    \PC(\name{MP}_i|\name{HBSS}_i)=\\
    \sum_j \PC(\name{MP}_i|\name{HBSS}_i,\name{PO}_{i,j})\PC(\name{PO}_{i,j}|\name{HBSS}_i)
\end{multline}
 
Note that if $\name{PO}_{i,j}$ and $\name{HBSS}_i$ are independent, we also have $\PC(\name{PO}_{i,j}|\name{HBSS}_i)=\PC(\name{PO}_{i,j})=\PC(\name{PO}_j)$. Although the template does not require this independence, $\PC(\name{PO}_j)$ can be reused for any $\name{HBSS}_i$ where the independence holds, simplifying the analysis. The independence often holds in practice, but not always. For example, it is reasonable to assume that the PO condition representing lighting conditions is independent from the HBSS of crossing an intersection as in the left-turn scenario defined by $\name{HBSS}_{\name{IL}}$. On the other hand, the probability of the PO condition of poor visibility due to precipitation is likely higher under the HBSS condition of reduced road friction. 

The sub-claims \clm{\name{PO}$_{i,j}$-MR} and \clm{\name{PO}$_{i,j}$} correspond to the terms in the summation in Eqn~\ref{eqn:po} and denote the conditioned misperception rate and occurrence of the PO condition, respectively. 
Thus, the relationship between bounds is: $\gamma_{i,\scriptsize{\name{MR}}}\ge \sum_j \gamma_{i,j,\scriptsize{\name{MR}}}\gamma_{i,j,\scriptsize{\name{PO}}\uparrow}$.

The effectiveness of this decomposition strategy depends on whether we can select a set $\{\name{PO}_{i,j}\}_j$ of conditions that well cover the categories of external triggers for hazardous misperceptions defined by \name{MP}$_i$.  
One way to achieve this is to base the set of PO conditions on a safety analysis method such as CV-HAZOP~\cite{kuwajima2019open,zendel2015cv}. CV-HAZOP defines a systematic method of exhaustively identifying modes of interference with a computer vision (CV) process by first modeling the CV process, and then using guide words in the spirit of HAZOP (Hazard and Operability Study) to identify how the process can be disturbed. The resultant list of interference modes can then be filtered and aggregated based on the specifics of the CV task~\cite{zendel2017analyzing}, ODD, HBSS and MP conditions to produce a specifically applicable PO set. A similar strategy could be used to analyze any perceptual task. 

\subsection{Goal \clm{\name{PO}$_{i,j}$} and Solution \sol{\name{PO}$_{i,j}$}}
The claim in this goal bounds the occurrence of ADS drives in the ODD that exhibit the condition \name{PO}$_{i,j}$. Here we require both upper and lower bounds to allow us to use the following for the nominal case: 
$$\PC(\name{PO}_{i,\scriptsize{\name{Nom}}})=1-\sum_{j\neq \scriptsize{\name{Nom}}}\PC(\name{PO}_{i,j})\le 1-\sum_{j \neq \scriptsize{\name{Nom}}}\gamma_{i,j,\scriptsize{\name{PO}}\downarrow}$$
As with \clm{\name{HBSS}$_i$}, it may be reasonable to assume that the occurrence of these conditions are independent of the specifics of the ADS design and choice of component \pc; thus, the bounds $\gamma_{i,j,\scriptsize{\name{PO}}\downarrow}$ and  $\gamma_{i,j,\scriptsize{\name{PO}}\uparrow}$ could be based on generic empirical or analytical sources of data. For example, information on precipitation frequency, visibility and sunlight variation may be sourced from weather bureaus, the frequency of different vehicle colors and shapes could come from vehicle sales statistics, etc. An important additional consideration is that the occurrence of these conditions may vary based on the geography in which the ADS will operate.

\subsection{Goal \clm{\name{PO}$_{i,j}$-MR}} 
This claim bounds the rate of $\name{MP}_i$ misperceptions within the context of a particular PO condition. 
Estimating the bound $\gamma_{i,j,\text{MR}}$ directly is difficult since \name{MP}$_i$ is a condition on \emph{sequences} of misperceptions, but component \pc~only operates at the single frame level producing individual misperceptions. Thus, there is a need to link these two levels of representation of misperceptions. The solution is to decompose the probability of \name{MP}$_i$ with respect to the probabilities of its constituent frame misperception patterns. 

\subsection{Strategy \stg{\name{PO}$_{i,j}$-fMP}}
The condition \name{MP}$_i$ is expressed in terms of a set $\{\name{fMP}_{i,k}\}_k$ of frame misperception patterns. For each $\name{fMP}_{i,k}$, a sub-claim \clm{\name{fMP}$_{i,j,k}$-MR} is created. In addition, we specify a \emph{linking expression} $F_{i,j}$, given in context node \ctx{Link$_{i,j}$}, that bounds the probability of \name{MP}$_i$ in terms of the probabilities of $\{\name{fMP}_{i,k}\}_k$.

Although the set $\{\name{fMP}_{i,k}\}_k$ is fixed for a given \name{MP}$_i$, the linking expression can vary depending on condition \name{PO}$_j$ because this can affect relationship between probabilities. For this reason, $F_{i,j}$  depends on both indices $i$ and $j$. For example, in the HMP \name{IL} the misperception pattern of repeated frame mis-localization events can cause an incorrect speed estimate of the on-coming vehicle, leading to a hazardous left turn. If a sequence of mis-localization events are caused by interference from precipitation, then each event can be considered to be independent in the context of precipitation PO condition. However, if the mis-localizations are caused by the reflectance characteristics of the on-coming car's surface (i.e., reflectance PO condition), then the events are not independent because the reflectance problem will continue to occur in all events.

\subsection{Goal \clm{\name{fMP}$_{i,j,k}$-MR}}
Since each $\name{fMP}_{i,k}$ is defined over individual frames, the claim for \clm{\name{fMP}$_{i,j,k}$-MR} uses the distribution $P_{C,\textit{fr}}$ of frames rather than the distribution \PC~of states in ADS drives. However, the probability is still conditional, restricted to frames occurring during drives satisfying \name{HBSS}$_i$ and \name{PO}$_i$. Since this goal involves testing or analysis at the unit level, it is marked as an away goal for \mdl{\pc-unit}. 

Note that different HMPs may share reliance on the same frame misperception patterns. For example, the fMP $\name{FN}_{20}$, representing FNs within 20 meters of the ego vehicle, may be used in definitions of different MP conditions. However, even if in \name{HMP}$_i$ and \name{HMP}$_{i'}$, $\name{fMP}_{i,k}=\name{fMP}_{i',k'}=\name{FN}_{20}$, the goals \clm{\name{fMP}$_{i,j,k}$-MR} and \clm{\name{fMP}$_{i',j',k'}$-MR} that bound its occurrence remain distinct because they depend on different HBSS and PO conditions. 

A key contribution of \plink~is that the set of claims $\{$\clm{\name{fMP}$_{i,j,k}$-MR}$\}_{i,j,k}$ can be seen to define a set of \emph{risk-aware performance metrics} for evaluating a component \pc~implementing task $T$:
\BD[Risk-aware Performance Metric]
Given test dataset \name{TDS}$_{i,j}=\{(x_l,y_l)\}_{l}$ drawn from conditional distribution $P_{C,\textit{fr}}((x,y)|\name{HBSS}_i,\name{PO}_{i,j})$, a \emph{risk-aware performance metric} $m_{i,j,k}$ for component $C$ is defined as:
$$m_{i,j,k}=\frac{1}{|\name{TDS}_{i,j}|}\sum_{(x,y)\in \name{TDS}_{i,j}}\mathbf{1}[C(x)\notin \name{fMP}_{i,k}(y)]  $$
\ED

Metric $m_{i,j,k}$ is a \emph{performance} metric because it computes a measure of the misperceptions produced by $C$. It is \emph{risk-aware} because it only counts hazardous misperceptions and ignores benign ones. Unlike ``generic'' performance metrics typically used for evaluating perception components  (e.g., recall, mAP, AuPR) that count any deviation from ground truth as bad, here only deviations that satisfy $\name{fMP}_{i,k}$ are considered bad. Furthermore, $\name{fMP}_{i,k}$ is derived from, and is traceable to, safety claims about $C$ via the argument in \plink. Finally, another benefit is that each metric $m_{i,j,k}$ focuses on a different system hazard and perception context (via HBSS $i$ and PO $j$) and a different aspect of the performance of $C$ (via fMP $k$), allowing a more fine-grained tuning of how $C$ impacts system safety.   

Goal \clm{\name{fMP}$_{i,j,k}$-MR} is a unit-level claim on the performance of component $C$; thus, it is expressed as an away-goal that links to module \mdl{$C$-unit}. While the full details of this argument are beyond the scope of this paper\footnote{See the recent related work discussed in the introduction for approaches to the argument in \mdl{$C$-unit}.}, the metric $m_{i,j,k}$ can be used as part of a black-box testing argument since, from a statistically standpoint, it is a sample estimate of  $P_{C,\textit{fr}}(\name{fMP}_{i,k}|\name{HBSS}_i,\name{PO}_{i,j})$. 

To use $m_{i,j,k}$ to compute $\gamma_{i,j,k}$ we need to take into account the sampling error of this estimate with sample size $N=|$\name{TDS}$_{i,j}|$. The sampling distribution is binomial but can be approximated with a Normal distribution when $N>30$. Then the upper bound $\sigma_q$ of the $q\%$ confidence interval on the error of $m_{i,j,k}$ is given by 
$$\sigma_q=z(q)\sqrt{\frac{m_{i,j,k}(1-m_{i,j,k})}{N}}$$ Where, $z(q)$ is the $\frac{100+q}{200}$ quantile of the standard normal distribution.  For example, if $q=99$ then $z(q)=2.58$. Then we can define $\gamma_{i,j,k}=m_{i,j,k}+\sigma_q$. This bound assumes that $\name{TDS}_{i,j}$ is a representative sample and data adequacy arguments in \mdl{$C$-unit} are applicable here. 
   

\section{Applying \plink~in Practice}\label{sec:discussion}
In this section, we discuss various topics related to the use of the \plink~during the development of \pc~and for constructing argument \mdl{\pc}.
\subsection{Accommodating Multiple Severity Levels}\label{sec:severity}
\plink~is designed for a single high severity event (i.e., crash); however, safety cases typically address multiple severity levels (e.g., four severity levels in ISO 26262). A simple and flexible way to do this is to create multiple \emph{parallel} arguments by adding an implicit parameter $L$ to the template representing $N$ severity levels. Then each GSN node can be interpreted as an array of $N$ nodes corresponding to the severity levels. For example, goal \clm{\name{PO}$_{i,j}$-MR}  is interpreted as $\PC(\name{MP}_i[L]|\name{HBSS}_i[L],\name{PO}_{i,j}[L])\le\gamma_{i,j,\scriptsize{\name{MR}}}[L]$ allowing the definitions of conditions \name{MP}$_i$, \name{HBSS}$_i$, \name{PO}$_{i,j}$ and bound $\gamma_{i,j,\scriptsize{\name{MR}}}$ to be qualified by severity level. When a condition or bound is not dependent on severity, the severity parameter need not be considered in the definition.

For example, in the HBSS for HMP \name{IL} (discussed in the preliminaries section), the severity of the hazardous behaviour of turning left when the on-coming vehicle is too close varies depending on the speed of the on-coming vehicle. Thus, \name{HBSS}$_{\name{IL}}[L]$ represents a version of the HBSS condition with a different speed depending on $L$. Correspondingly, in \name{MP}$_{\name{IL}}[L]$ the number of required $\name{FN}_{20}$ misperceptions may reduce with increasing speed. However, if \name{MP}$_{\name{IL}}$ represents misperceptions of a direct speed measurement (e.g., via Radar) then it would not be dependent on $L$.

\subsection{Top-down vs. Bottom-up Development}\label{sec:use}
\plink~takes a formal deductive approach by using the decompositional structure of the argument to express bound $\gamma_{\scriptsize{{\pc}}}$ in the top-level claim in terms of similar bounds in lower level claims. The mathematical relationship between the bounds given in each strategy defines formal traceability from every lower-level claim to the top-level claim. 

The formal traceability allows $\gamma_{\scriptsize{{\pc}}}$ to be interpreted in either of two ways: i) as a \emph{safety target} that component \pc~must achieve, or, ii) as a conservative (i.e, upper bounding) estimate of the probability \PC$(\name{MCrash})$ based on its sub-claims. Interpretation (i) supports a top-down development strategy in which a safety target from the main ADS argument \mdl{ADS} is systematically allocated to different sub-claims to define component level requirements for \mdl{\pc-unit}. Interpretation (ii) supports a bottom-up development strategy in which confidence
about leaf claims (as expressed by the bounds on these) are correctly propagated upward to \mdl{ADS}.  
This can be used to guide development effort on the component by identifying which HMPs, fMPs, and PO conditions contribute the most to risk.

\subsection{Iterative and Continuous Development}
Developing a safety case is costly, thus any iteration regarding a component that impacts its safety case must incur a cost. However, ML-based perception components may be frequently re-trained as new useful training samples become identified or when domain shift  occurs. In addition, some development methods, such as active learning, require multiple retraining. 

\plink~has the beneficial property that it largely independent of the particular choice of component \pc. The structure (i.e., choice of claims) is determined by perception task $T$ and is independent of \pc. The only claims that are directly affected by \pc~are \clm{\name{fMP}$_{i,j,k}$-MR} because the bounds $\gamma_{i,j,k}$ must be recomputed or rechecked when \pc~changes (although, the test datasets $\name{TDS}_{i,j}$ are independent of \pc). The bounds for higher claims can be automatically recomputed from these. Thus, the impact of iterating \pc~is well localized to limit the safety case change cost.
\section{Case Study}\label{sec:casestudy}
In this section, we demonstrate the instantiation of \plink~for an object detection task \name{OD} and define the claims associated with one HMP in detail---HMP \SCA~(``stopped car ahead'').  The component implementing the task \name{OD} is \SSD, the Point Pillars object detector for LiDAR point clouds \cite{lang2019pointpillars}.
Knowing the internal details of \SSD~is not needed for understanding this case study. 

The ODD we assume consists of driving conditions represented in the KITTI object detection dataset~\cite{geiger2012we}, which is naturalistic, summer, clear weather, day-time  driving in a small city (based on Karlsruhe, Germany). 

The ADS we used in which \SSD~operates 
has the following details relevant to the case study:
\begin{itemize}
    \item In the perception pipeline, the output of \SSD~feeds a \emph{Tracker} component that maintains a model of past and predicted (near future) trajectories of all relevant road users. It takes $n_{trk}=9$ consecutively missed frames by \SSD~for tracker to lose the track of an object.
    \item The comfortable and maximum (i.e., emergency) braking rates capable by the ego vehicle are $a_{b,comf}=2.01\, m/s^2$ and $a_{b,emerg}=2.86\,m/s^2$, respectively. The maximum acceleration is $a_{max}=3.02\, m/s^2$. Maximum speed is $11.11\,m/s$ ($40\,km/h$).
       \item The frame rate for task \name{OD} is $10\,f/s$
\end{itemize}
  
We use FOL to sketch formal definitions throughout this section.

\subsection{Goal \clm{\SSD}}
The task \name{OD} is defined by ground truth function $\name{OD}:X\to Y$ where $X$ are LiDAR point clouds and $Y$ are sets 
of bounding boxes classified as car, pedestrian or cyclist. Component 
\SSD~implementing \name{OD} defines function $\SSD:X\to Y$ and may deviate from \name{OD} resulting in FNs and FPs. Fig.~\ref{fig:misper} shows 
two examples in the context of a hazardous frame misperception pattern \FNA~defined below. The objective of the claim in this goal is to show that the occurrence of hazardous misperceptions by \SSD~is bounded to an acceptable level $\gamma_{\ssd}$ as determined by the main ADS argument \mdl{ADS}. 
\subsection{Strategy \stg{HMP}}
In this example, we are only considering the HMP \SCA. 
\subsection{Goal \clm{\name{HMP}$_{\sca}$}}\label{sec:hmpsca}
We assume that, in the hazard analysis used by the main ADS argument, the following hazardous operational situation is identified: the ego vehicle, gets close enough to a stopped car ahead, such that, unless braking is applied by the ego vehicle a crash (i.e., collision with stopped vehicle) will result. Thus, any \emph{braking interruption} of sufficient time length to cause an accident is a hazardous behaviour of the ADS in this situation. 

This system-level hazard could have different causes including slippery roads and malfunctioning brakes but since a misperception in performing \name{OD} can also be a cause, we create the HMP \SCA $=\tuple{\name{HBSS}_{\sca}, \name{MP}_{\sca}}$ to represent this case within argument \mdl{\SSD}.

To define \name{HBSS}$_{\sca}$, we  assume the following safe driving policy: a stopped vehicle should be detected sufficiently far ahead to allow the ego vehicle, braking comfortably, to stop at a stand-still distance of 4m from it. In addition, based on simple physics-based modeling we observe that if the ego vehicle is travelling at speed $v$, it must brake with at least $a_b$ to avoid a collision with a stopped vehicle distance $x = \frac{v^2}{2a_{b}}$ ahead.
\BD [$\name{HBSS}_{\sca}(d)$]
For all drives $d\in \mathcal{D}$, $\name{HBSS}_{\sca}(d)$ iff there is a stopped vehicle $x_{sc} = \frac{v^2_{init}}{2a_{b,comf}}+4$ meters ahead of the ego vehicle and the ego speed in state $d[1]$ is $v_{init}>0$.
\ED


Thus, $\name{HBSS}_{\sca}(d)$ then, under normal operation, the ego vehicle stops safely. However, if there is a braking interruption, it will need to compensate by braking more intensely (up to a maximum of $a_{b,emerg}$) once braking resumes. The braking interruption becomes a hazardous behaviour if at any point, $x < \frac{v^2}{2a_{b,emerg}}$, where $v$ is the ego vehicle speed and $x$ is the remaining distance to the stopped vehicle, since a collision will result. 

Different misperceptions could lead the ADS to interrupt braking, including: not detecting the stopped car, misjudging the position of the stopped car as not obstructing the ego vehicle, and misjudging the speed of the car ahead as not stopped.
In this study, we focus on the first of these---missed detections (i.e., FNs)---and assume that if the car is detected, other information about it will be correctly perceived. 

To define $\name{MP}_{\sca}$, we first define the frame misperception pattern $\FNA : Y \to pow(Y)$ identifying an FN for the vehicle ahead.
\BD[Frame misperception pattern \FNA]
\begin{multline*}
    \forall y, y_{\scriptsize{\name{gt}}}\in Y\cdot y\in\FNA (y_{\scriptsize{\name{gt}}}) \text{ iff } \\ 
    (\exists bb_{\scriptsize{\name{gt}}}\in y_{\scriptsize{\name{gt}}} \cdot  \name{AheadOf}(y_{\scriptsize{\name{gt}}}, bb_{\scriptsize{\name{gt}}},\name{Ego}))\wedge \\
    (\neg \exists bb\in y \cdot \name{TPMatch}(bb_{\scriptsize{\name{gt}}},bb))
\end{multline*}
Where, $\name{AheadOf}(y,bb,bb')$ iff bounding box $bb\in y$ is positioned ahead of $bb'$ with no other in between, \name{Ego} is the bounding box for the ego vehicle and $\name{TPMatch}(bb,bb')$ iff $bb'$ matches $bb$ well enough to be considered a true positive prediction for $bb$.
\ED

\begin{figure}
    \centering
    \includegraphics[width=0.40\textwidth]{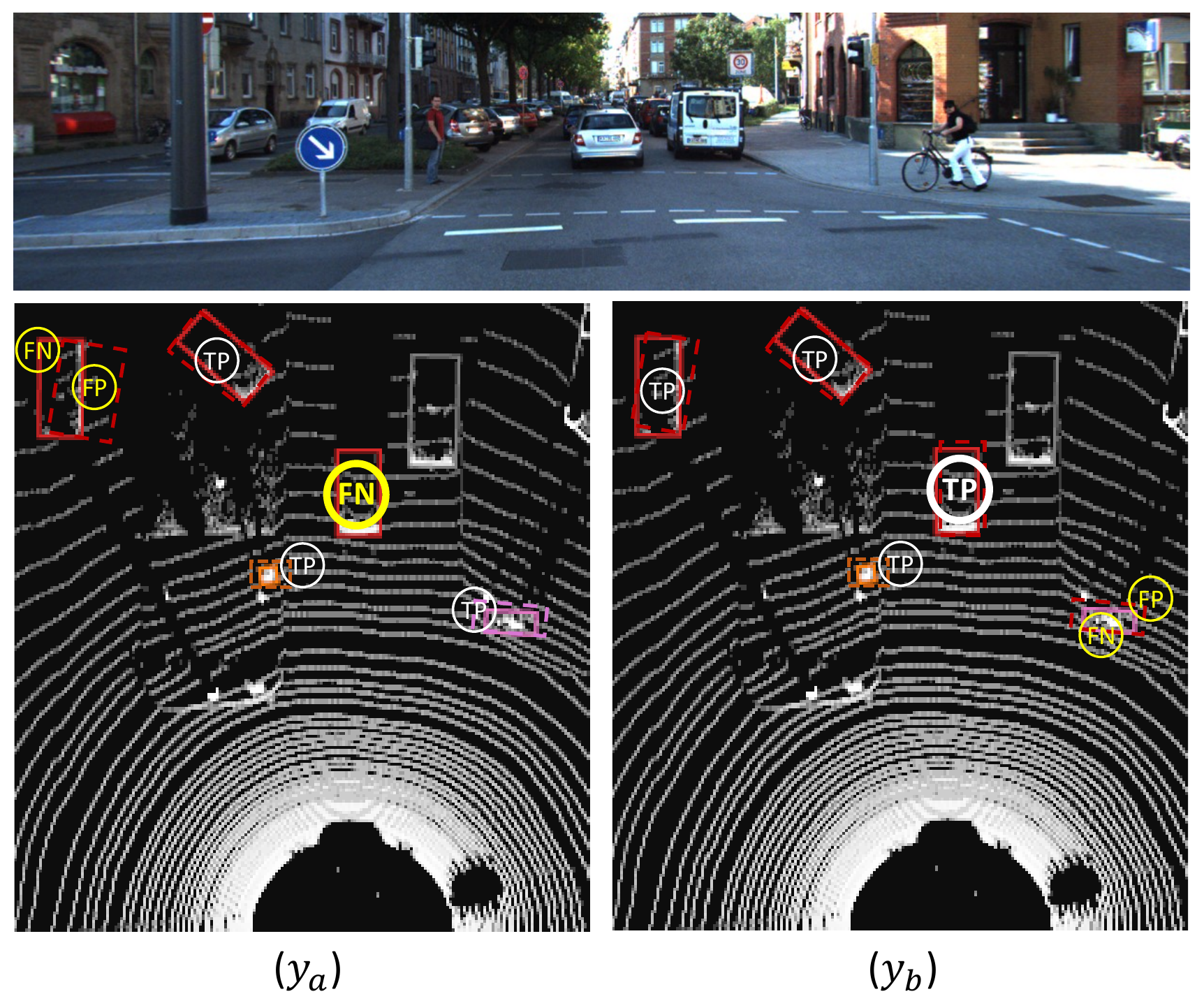}
    \caption{
    Illustration of frame misperception pattern \FNA~on two point cloud outputs of task \name{OD} with same input $x$ (predictions are dashed boxes, ground truth are solid boxes).  In output
    $y_a$, \FNA~occurs since the vehicle ahead of ego
    vehicle is FN (shown larger), thus $y_a\in\FNA (\name{OD}(x))$. In output $y_b$,
    the vehicle ahead is TP, thus \FNA~does not occur and
    $y_b\notin\FNA (\name{OD}(x))$. Accuracy on other objects is irrelevant for \FNA. Classes: Car (red), Pedestrian (orange), Cyclist (purple), Other (gray).}
    \label{fig:misper}
\end{figure}

Fig.~\ref{fig:misper} illustrates \FNA. To define   $\name{MP}_{\sca}$ we must determine what sequences of \FNA~occurrences yield enough a braking interruption to be hazardous. To do this, we conducted a physics-based analysis combined with ADS simulation and determined that the worst-case (minimum) hazardous braking interruption is a single interruption of $t_{crash}=0.48\,s$ beginning $19.65\,m$ from the stopped vehicle with $v_{init}=11.11m/s$ (i.e., speed limit) and an ego vehicle acceleration of $a_{max}=3.02\,m/s^2$ during the interruption. Given the frame rate of $10\,f/s$, this means that braking must be interrupted for $5$ frames. However, since the tracker component requires $n_{trk}=9$  consecutive FNs before the track is lost, \SSD~must exhibit at least $n_{crash}=5+9=14$  \FNA~occurrences to cause a hazardous braking interruption. 
\BD[$\name{MP}_{\sca}$]\label{def:MPsca}
For all drives $d\in \mathcal{D}$, $\name{MP}_{\sca}(d)$ iff it contains at least $n_{crash}=14$, not necessarily consecutive, occurrences of \FNA
\ED

Note that consecutiveness is not specified here because there are nonconsecutive sequences of \FNA s that can be hazardous, but based on the above analysis, more than $14$ \FNA s are required.

\subsection{Goal \clm{Res} and Solution \sol{Res}}
Since we are only considering a single HMP in this case study, this goal is not applicable.  

\subsection{Goal \clm{\name{MP}$_{\sca}$-N} and Solution \sol{\name{MP}$_{\sca}$-N}}
Evidence for this claim is based on the argument that since we are limiting the focus to braking interruptions due to \name{OD} misperceptions, this could only have been caused by occurrences of \FNA~misleading the ADS into believing there is no car ahead. Furthermore, by the physics/simulation analysis discussed in Sec.~\ref{sec:hmpsca}, producing a hazardous braking interruption requires at least $14$ \FNA s. Finally, by Def.~\ref{def:misperc}, $\name{MP}_{\sca}$ represents all possible sequences containing at least $14$ \FNA s. Therefore, we conclude  $P_{\ssd}(\name{MCrash}|\name{HBSS}_{\sca},\neg\name{MP}_{\sca})\cong 0$ (approximation to account for a potential small error in analysis).

\subsection{Goal \clm{\name{MP}$_{\sca}$-CR} and Solution \sol{\name{MP}$_{\sca}$-CR}}
The physics/simulation analysis says that fewer than $14$ \FNA s cannot produce the hazardous behaviour, but it is still possible that $14$ or more nonconsecutive \FNA s may not be hazardous. For example, having two nonconsecutive groups of nine \FNA s is nonhazardous due to the tracker compensation. Although a simulation study could yield a more accurate estimate of this crash rate, for this claim we take a conservative stance and set $\gamma_{\sca,\scriptsize{\name{CR}}}=1$.




\subsection{Goal \clm{\name{HBSS}$_{\sca}$} and Solution \sol{\name{HBSS}$_{\sca}$}}
We assume that the bound in this claim is based on information from the corresponding system-level hazard in \mdl{ADS}. For this case study, we conservatively assume that $\name{HBSS}_{\sca}$ occurs at every intersection. If the ego vehicle is travelling at the speed limit ($11.11\,m/s$) and the average distance between intersections is $500\,m$, then we let $\gamma_{\sca,\scriptsize{\name{HBSS}}}=1.11/500 = 0.0022$ since $1.11\,m$ is travelled per state and one state ends an $\name{HBSS}_{\sca}$ drive per $500\,m$.

\subsection{Goal \clm{\name{MP}$_{\sca}$-MR} and Strategy \stg{\name{MP}$_{\sca}$-PO}}
The claim is $P_{\ssd} (\name{MP}_{\sca}|\name{HBSS}_{\sca})\leq \gamma_{\sca,\scriptsize{\name{MR}}}$. To decompose HMP \SCA~we consider a single non-nominal PO condition: $\name{PO}_{\pob}$ identifying crowded scenes defined as those containing more than 10 objects within 40\,m of the ego vehicle. This is based on the assumption that crowded scenes will contain more FNs than uncrowded.

\subsection{Goals \clm{\name{PO}$_{\sca,j}$} and Solutions \sol{\name{PO}$_{\sca,j}$}}
We estimate the probability of crowded scenes by the proportion of such scenes in the KITTI test dataset and use the $99\%$ confidence interval to account for sampling error. The bound estimates are given in Table~\ref{tab:po}.

\begin{table}
        \centering
        \caption{Computation of bounds $\gamma_{\sca,j,\scriptsize{\name{PO}}\downarrow}$ and $\gamma_{\sca,j,\scriptsize{\name{PO}}\uparrow}$.}
        \begin{tabular}{|c|c|c|c|c|}
             \hline
             $j$ & $\frac{|\name{TDS}_{j}|}{\name{|TDS|}}$ & $\sigma_{0.99}$ &  $\gamma_{\sca,j,\scriptsize{\name{PO}}\downarrow}$ &$\gamma_{\sca,j,\scriptsize{\name{PO}}\uparrow}$ \\
             \hline
             \poa & 0.965 & 0.008 & 0.957 & 0.973 \\
             \hline
             \pob & 0.035 & 0.008 & 0.027 & 0.043  \\
             \hline
        \end{tabular}
    \label{tab:po}
\end{table}

\subsection{Goals \clm{\name{PO}$_{\sca,j}$-MR} and Strategies \stg{\name{PO}$_{\sca,j}$-fMP}}
We decompose each goal \clm{\name{PO}$_{\sca,j}$-MR} using the frame misperception pattern \FNA. 
If we assume each \FNA~occurrence is independent, then given Def.~\ref{def:MPsca}, we have the following expression as expected value of the cumulative binomial over drives of varying length $n$,
\begin{equation*}
\begin{split}
    P_{\ssd}(\name{MP}_{\sca}|\name{HBSS}_{\sca}, \name{PO}_{\sca,j}) &=\\  & \mathbb{E}_n\left[{\sum_{l=14}^{n}{{n\choose l} p_j^l (1-p_j)^{(n-l)}}}\right]
\end{split}
\end{equation*}

Where, $p_j = P_{\ssd,fr}(\FNA|\name{HBSS}_{\sca}, \name{PO}_{\sca,j})$. Note that the cumulative binomial is monotonic in $n$ and max $n=55$ occurs when starting with max $v_{init}=11.11m/s$. Thus, for the linking expression we use:
\begin{equation*}
\begin{split}
    P_{\ssd}(\name{MP}_{\sca}|\name{HBSS}_{\sca}, \name{PO}_{\sca,j}) & \leq  \\
    & \sum_{l=14}^{55}{{55\choose l} p_j^l (1-p_j)^{(55-l)}}
\end{split}
\end{equation*}

\subsection{Goals \clm{\name{fMP}$_{\sca,j,\fna}$-MR}}
The objective here is to estimate $P_{\ssd}(\FNA|\name{HBSS}_{\sca}, \name{PO}_{\sca,j})$. To do this, we extracted frames from the KITTI dataset conforming to \name{HBSS}$_{\sca}$ and \name{PO}$_{\sca,j}$ to form datasets \name{TDS}$_{\sca,j}$ and used these to test \SSD~and compute the risk-aware metrics $m_{\sca,j,\fna}$. Specifically, \name{TDS}$_{\sca,j}$ consisted of frames in which there was a car ahead of the ego vehicle. These frames over-approximate the set of frames from  \name{HBSS}$_{\sca}$ drives because it doesn't consider the speed of the ego vehicle ($v_{init}$) or whether the car ahead is stopped or moving; however, the single-frame misperception performance is unaffected by this. Tables~\ref{tab:fna} and~\ref{tab:fna1} give the results assuming a $99\%$ confidence bound. 




 
\begin{table}
    \centering
    \caption{Computation of metric  $m_{\sca,j,\fna}$.}        \begin{tabular}{|c|c|c|}
             \hline
             $j$ & $|\name{TDS}_{\sca,j}|$ & $m_{\sca,j,\fna}$  \\
             \hline
             \poa & 1479 & 0.052 \\
             \hline
             \pob & 67 & 0.045 \\
             \hline
        \end{tabular}
        \vspace{0.1in}

    \label{tab:fna}
\end{table}

\begin{table}
    \centering
    \caption{Computation of bounds $\gamma_{\sca,j,\fna}$ and $\gamma_{\sca,j,\scriptsize{\name{MR}}}$.}        \begin{tabular}{|c|c|c|c|}
             \hline
             $j$ & $\sigma_{0.99}$ & $\gamma_{\sca,j,\fna}$ & $\gamma_{\sca,j,\scriptsize{\name{MR}}}$ \\
             \hline
             \poa & 0.015 & 0.067 &$1.15\times 10^{-5}$ \\
             \hline
             \pob & 0.065 & 0.110 & $2.06\times 10^{-3}$\\
             \hline
        \end{tabular}
        \vspace{0.1in}

    \label{tab:fna1}
\end{table}

\subsection{Combining the Case Study Results}
By propagating the values determined for the bounds in the leaf claims upward using the expressions in the strategies (See Fig.~\ref{fig:sct}), we obtain bound values for higher level claims. 
Based on the values in Table~\ref{tab:po} and Table~\ref{tab:fna1} for PO conditions \name{Nom} and \name{Crowd}, we have:
\begin{equation*}
\begin{split}
\gamma_{\sca,\scriptsize{\name{MR}}}&=
\gamma_{\sca,\poa,\scriptsize{\name{PO}}\uparrow}\gamma_{\sca,\poa,\scriptsize{\name{MR}}}+\gamma_{\sca,\pob,\scriptsize{\name{PO}}\uparrow}\gamma_{\sca,\pob,\scriptsize{\name{MR}}} \\
&=(0.973)(1.15\times 10^{-5})+(0.043)(2.06\times 10^{-3}) \\
&=1.12\times 10^{-5}+8.86\times 10^{-5} \\
&=9.98\times 10^{-5}
\end{split}
\end{equation*}

We can see that the dominant contribution is from PO condition \name{Crowd}. The bound for \clm{HMP$_{\sca}$} can then be computed as:
\begin{equation*}
\begin{split}
\gamma_{\sca}&=\gamma_{\sca,\scriptsize{\name{CR}}}\gamma_{\sca,\scriptsize{\name{MR}}}\gamma_{\sca,\scriptsize{\name{HBSS}}} \\
&=(1)(9.98\times 10^{-5})(2.2\times10^{-3})=2.20\times10^{-7}
\end{split}
\end{equation*}

\noindent
Finally, the bound for the top claim is:
$$\gamma_{\ssd}=\gamma_{res}+\gamma_{\sca}=\gamma_{res}+2.20\times10^{-7}$$
Since this analysis is based on only one HMP, it is partial, and we leave the residual bound $\gamma_{res}$ as an unspecified variable term.










 

\section{Summary/Conclusions} \label{sec:conclusion}
In this paper, we address a gap in the research on safety cases for automated driving and ML by proposing the \plink~ template safety argument linking the system-level arguments and unit-level arguments. The template provides a formally deductive claim decomposition approach tailored to perception and identifies a set of risk-aware safety metrics that can be used to evaluate perception components. We demonstrate the applicability of \plink~through a detailed case study.

As part of future work, we explore several directions. First, the issue of \emph{confidence} needs special consideration. It is well known that the strength of an argument depends on the level of confidence in claims generated by the evidence and there is much research on defining and propagating confidence within an argument. We have addressed this using confidence intervals in some claims, but it requires a more systematic treatment throughout the argument in \plink. 
Second, while an underlying methodology for identifying HMPs and PO conditions is suggested in various claims, this needs comprehensive elaboration. 

Third, we intend to extend \plink~to address some common variations, including the following:
\begin{itemize}
    \item Currently, only a single linking expression is allowed for connecting MP conditions to their constituent frame misperception patterns but it is clear that the expression could also depend on other factors. For example, in the case study, we assume independence between occurrences of \FNA s. This is appropriate for ``typical'' cars, but for cars that are unusual it is likely that if there is one \FNA, then all subsequent detections will be \FNA, so the independence assumption is not valid. Thus, having different linking expressions for typical and atypical cars is needed here.
    \item Currently, HBSS exposure is assumed to be frequency-based, which is appropriate for cases such as when stopping for a stopped vehicle. All uses of HBSS exposure in \plink~are based on this assumption. However, ISO2626 also allows for duration-based HBSS exposure, such as when following a vehicle at a steady pace. The template should allow either approach for generality.
    \item Currently, the frame rates of input and output of perception task $T$ are assumed to be the same, but in general, they do not have to match. For example, a multi-rate tracker might use 30Hz camera input, a 10Hz LiDAR input and output at 20Hz. The template should be generalized to accommodate such cases.
    \item Currently, PO conditions are assumed to represent special (non-nominal) cases and the distinguished PO condition \name{Nom} captures all remaining (assumed to be nominal) cases. For greater generality, the \name{Nom} PO condition should be split to further distinguish nominal cases from any residual non-nominal cases that may exist.

\end{itemize}

Finally, we intend to do a detailed feasibility study of the approach and identify places it can be improved with the hope that eventually \plink~can be adopted to support industrial practice.

\bibliographystyle{IEEEtran}
\bibliography{references}

\section*{Acknowledgments}
The University of Waterloo authors were supported, in part, by DENSO CORPORATION.

\end{document}